\documentclass[10pt,twocolumn,letterpaper]{article}

\usepackage[pagenumbers]{iccv} 

\definecolor{iccvblue}{rgb}{0.21,0.49,0.74}
\usepackage[pagebackref,breaklinks,colorlinks,allcolors=iccvblue]{hyperref}

\def\paperID{9420} 
\def\confName{ICCV}
\def\confYear{2025}

\title{VIVID-10M: A Dataset and Baseline for Versatile and Interactive \\Video Local Editing}

\author{Jiahao Hu$^{1,*}$, Tianxiong Zhong$^{2,*}$, Xuebo Wang$^3$, Boyuan Jiang$^3$, Xingye Tian$^{3,}$\footnotemark[2]\ , \\
Fei Yang$^3$, Pengfei Wan$^3$, Di Zhang$^3$\\
$^1$Northwest Polytechnical University Xi'an, $^2$Beijing Institute of Technology, $^3$Kuaishou Technology\\
{\tt\small noreasonhjh@mail.nwpu.edu.cn, inkosizhong@gmail.com,} \\
{\tt\small \{wangxuebo,jiangboyuan,tianxingye,yangfei06,wanpengfei,zhangdi08\}@kuaishou.com}
 \vspace{-0.4cm}
}

\begin{document}
\twocolumn[{%
\renewcommand\twocolumn[1][]{#1}%
\maketitle
\begin{center}
    \centering
    \includegraphics[width=\linewidth]{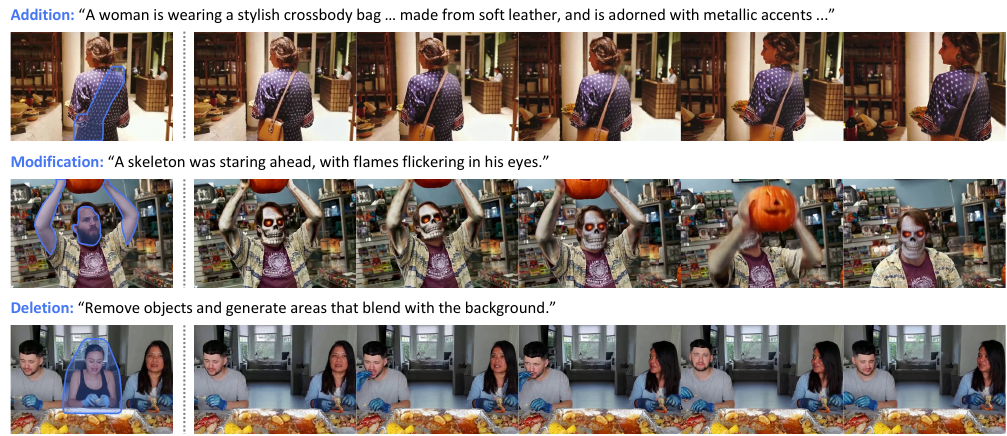}
    \captionof{figure}{Our method enables seamless addition, modification, and deletion of entities in videos. Edits are guided by masks and text, which specify both the target position and desired content.}
    \label{fig:teaser}
\end{center}%
}]
\begin{abstract}
Diffusion-based image editing models have made remarkable progress in recent years.
However, achieving high-quality video editing remains a significant challenge.
One major hurdle is the absence of open-source, large-scale video editing datasets based on real-world data, as constructing such datasets is both time-consuming and costly.
Moreover, video data requires a significantly larger number of tokens for representation, which substantially increases the training costs for video editing models.
Lastly, current video editing models offer limited interactivity, often making it difficult for users to express their editing requirements effectively in a single attempt.
To address these challenges, this paper introduces a dataset VIVID-10M and a baseline model VIVID.
VIVID-10M is the first large-scale hybrid image-video local editing dataset aimed at reducing data construction and model training costs, which comprises 9.7M samples that encompass a wide range of video editing tasks.
VIVID is a Versatile and Interactive VIdeo local eDiting model trained on VIVID-10M, which supports entity addition, modification, and deletion. 
At its core, a keyframe-guided interactive video editing mechanism is proposed, enabling users to iteratively edit keyframes and propagate it to other frames, thereby reducing latency in achieving desired outcomes.
Extensive experimental evaluations show that our approach achieves state-of-the-art performance in video local editing, surpassing baseline methods in both automated metrics and user studies. 
The VIVID-10M dataset are open-sourced at \url{https://kwaivgi.github.io/VIVID/}.
\end{abstract}    

\definecolor{darkgreen}{RGB}{1,150,32}
\definecolor{darkred}{RGB}{150,32,1}

\begin{figure*}[htbp]
    \centering
    \begin{minipage}{0.6\textwidth}
    \setlength{\tabcolsep}{3pt} 
    \makeatletter\def\@captype{table}
        \centering
        \small
        \begin{tabular}{lcccccc}
        \toprule
        \multirow{2}{*}{Datasets} & Image & Video & Real & Editing & \multirow{2}{*}{Resolution} & \multirow{2}{*}{\#Records} \\
        & Edit & Edit & Data & Region & & \\
        \midrule
        InstructPix2Pix~\cite{brooks2023instructpix2pix} & \textcolor{darkgreen}\Checkmark &\textcolor{darkred}\XSolidBrush&\textcolor{darkred}\XSolidBrush& \textcolor{darkred}\XSolidBrush & 512$\times$512 & 313,010 \\
        HQ-Edit~\cite{hui2024hq} & \textcolor{darkgreen}\Checkmark &\textcolor{darkred}\XSolidBrush &\textcolor{darkred}\XSolidBrush &\textcolor{darkred}\XSolidBrush & 900$\times$900 & 197,350 \\
        UltraEdit-free~\cite{zhao2024ultraedit} & \textcolor{darkgreen}\Checkmark &\textcolor{darkred}\XSolidBrush& \textcolor{darkgreen}\Checkmark &\textcolor{darkred}\XSolidBrush& 512$\times$512 & 4,000,083 \\
        MagicBrush~\cite{zhang2024magicbrush} & \textcolor{darkgreen}\Checkmark &\textcolor{darkred}\XSolidBrush& \textcolor{darkgreen}\Checkmark & \textcolor{darkgreen}\Checkmark & 512$\times$512 & 10,388 \\
        UltraEdit-region~\cite{zhao2024ultraedit} & \textcolor{darkgreen}\Checkmark &\textcolor{darkred}\XSolidBrush& \textcolor{darkgreen}\Checkmark & \textcolor{darkgreen}\Checkmark & 512$\times$512 & 108,179 \\
        \midrule
        InsV2V~\cite{cheng2023consistent} &\textcolor{darkred}\XSolidBrush& \textcolor{darkgreen}\Checkmark &\textcolor{darkred}\XSolidBrush&\textcolor{darkred}\XSolidBrush& 256$\times$256 & 404,168 \\
        \textbf{VIVID-10M} & \textcolor{darkgreen}\Checkmark & \textcolor{darkgreen}\Checkmark & \textcolor{darkgreen}\Checkmark & \textcolor{darkgreen}\Checkmark & \textbf{1280$\times$720} & \textbf{9,768,279} \\
        \bottomrule
        \end{tabular}
        \caption{Compare VIVID-10M with existing image and video editing datasets.}
        \label{tab:dataset_compare}
    \end{minipage} \hfill
    \begin{minipage}{0.38\textwidth}
        \centering
        \includegraphics[width=\linewidth]{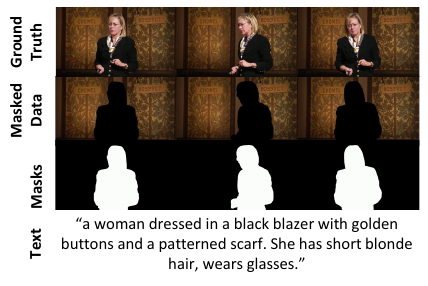}
        \caption{Each sample in VIVID-10M contains ground truth, masked data, masks and local caption.}
        \label{fig:dataset_vis}
    \end{minipage}
\end{figure*}
\section{Introduction}
\label{sec:intro}

Image and video editing based on diffusion models~\cite{DDPM,DDIM,song2020score} have achieved great progress in recent years.
Video editing algorithms, which generate edits based on a reference video and a provided description, can generally be classified into two categories: training-free~\cite{qi2023fatezero,cong2023flatten,kara2024rave,bai2024uniedit} and training-based~\cite{wu2023tuneavideo,tu2024motioneditor,qin2024instructvid2vid,cheng2023consistent,wang2024videocomposer,liew2023magicedit,zhang2024avid,zi2024cococo,polyak2024moviegen}.
Training-based algorithms typically achieve superior text alignment and temporal consistency.
To enable more precise and controllable video edits, local editing methods~\cite{liew2023magicedit,zhang2024avid,zi2024cococo} utilize mask sequences to define the editing regions, enhancing the ability to preserve background, \textit{i.e.}, maintaining non-editable areas unchanged.

However, achieving high-performance video local editing faces several challenges.
\textbf{C1. Lack of large-scale video editing datasets.}
Training-based algorithms require extensive high-quality paired data.
Some algorithms~\cite{qin2024instructvid2vid,cheng2023consistent} leverage large language models and training-free approaches to construct synthetic video datasets.
However, this approach is unable to generate local editing data, thereby constraining the performance of training-based models to the limitations of training-free approaches.
Video local editing algorithms~\cite{zhang2024avid,zi2024cococo} extract mask sequences from video frames via visual perception algorithms~\cite{zhang2024ram,liu2023groundingdino,kirillov2023SAM} and mask the original videos to generate paired data.
Despite using high-quality real-world video data, there is still no open-source large-scale dataset for video local editing tasks. Constructing such a dataset is challenging due to the time- and resource-intensive demands of the data processing pipeline.
\textbf{C2. High training overhead.}
Video editing models typically add temporal attention layers~\cite{zhang2024avid,zi2024cococo,wang2024videocomposer,liew2023magicedit} to image editing~\cite{brooks2023instructpix2pix} or generation models~\cite{rombach2022LDM}.
Video data also require more tokens to represent than image data, reducing the training efficiency of video editing models compared to image editing models.
\textbf{C3. Limited interactivity.}
Users often find it challenging to represent their editing requirements in a single attempt. 
This necessitates iterative adjustments and feedback cycles to refine the edits, leading to prolonged inference times during the video editing process. 
This lack of seamless interactivity prolongs the time to achieve desired results.


We address challenges C1 and C2 by leveraging a large volume of easily constructed image data to optimize the model's spatial modeling capabilities, while using video data to enhance spatio-temporal modeling.
To this end, we introduce VIVID-10M, a high-quality video local editing dataset, consisting of 9.7M samples derived from 73.7K videos and 672.7K images. 
Each video and image meets a resolution above 720p, with video clips spanning at least 5 seconds in duration.
VIVID-10M is constructed through an automated pipeline that cascades various visual perception models~\cite{liu2023groundingdino,zhang2024ram,SAM2,teed2020raft} and a multi-modality large language model~\cite{chen2024internvl}. 
Each sample includes ground truth, masks, masked data and local captions for addition, deletion, and modification tasks.
To evaluate VIVID-10M, we propose VIVID, a versatile and interactive video local editing model that supports entity addition, deletion, and modification (\cref{fig:teaser}). 
VIVID is jointly trained on image and video data to reduce training overhead, achieving state-of-the-art performance compared to existing methods~\cite{zi2024cococo,wang2024videocomposer,zhou2023propainter,li2025diffueraser}.

To address challenge C3, we propose a Keyframe-guided Interactive Video Editing mechanism (KIVE), enabling users to quickly achieve editing results for keyframes using an image editing model and propagate satisfactory results to the remaining frames. 
Additionally, since VIVID employs mixed image and video training, it is also applicable during the keyframe editing phase.
Experiments demonstrate that the KIVE mechanism significantly enhances user interactivity, leading to more efficient workflows and high-quality video editing outcomes. 
Furthermore, the KIVE mechanism supports local editing of long videos by using the last frame of one edited clip as the keyframe for the next.

In summary, we highlight the main contributions:
\begin{enumerate}
  \item We introduce VIVID-10M, the first large-scale high-quality dataset for video local editing.
  \item We present VIVID, a robust video local editing model that supports entity addition, modification, and deletion.
  \item We propose a Keyframe-guided Interactive Video Editing (KIVE) mechanism that enhances user experience by enabling iterative keyframe edits.
\end{enumerate}
\section{Related Work}\label{sec:related}
\subsection{Image and Video Editing Datasets}
Open-source image editing datasets have significantly contributed to advancing image editing models~\cite{brooks2023instructpix2pix,hui2024hq,zhang2024magicbrush,zhao2024ultraedit,ju2024brushnet,zhuang2023powerpaint,gal2022image_worth_word,han2024proxedit,kawar2023imagic,miyake2023negative}.
\cref{tab:dataset_compare} summarizes the existing image and video editing datasets. 
For instance, InstructPix2Pix~\cite{brooks2023instructpix2pix} and HQ-Edit~\cite{hui2024hq} use large language models (LLMs) to generate paired captions and editing instructions, with image generation models creating the corresponding images.
MagicBrush~\cite{zhang2024magicbrush} relies on human annotators to manually label data from image generation models.
UltraEdit~\cite{zhao2024ultraedit} uses the prompt-to-prompt~\cite{hertz2022prompt2prompt} mechanism and a modified image inpainting pipeline to generate free-form and region-based (local) editing samples, respectively.
In contrast, only one public video editing dataset, InsV2V~\cite{qin2024instructvid2vid}, is currently available, and it does not support local editing.
InsV2V synthesizes videos based on the captions generated by the LLM, and produces corresponding editing data through the prompt-to-prompt~\cite{hertz2022prompt2prompt} mechanism.
The lack of large-scale high-quality video editing datasets is a primary obstacle to the advancement of video editing.

\subsection{Training-free Video Editing}
Training-free video editing algorithms use pretrained image or video generation models~\cite{qi2023fatezero,cong2023flatten,kara2024rave,bai2024uniedit} to implement video editing in a training-free manner.
These algorithms apply DDIM inversion~\cite{song2020ddiminv} and incorporate additional mechanisms to ensure controllable, continuous and stable video editing.
For example, FateZero~\cite{qi2023fatezero} blends the self-attention maps with masks to stabilize non-editing areas.
FLATTEN~\cite{cong2023flatten} extracts inter-frame optical flow to guide self-attention calculations and improve temporal consistency.
RAVE~\cite{kara2024rave} shuffles latents across frames and concatenates them together as a large image for denoising to ensure temporal consistency.
UniEdit~\cite{bai2024uniedit} maintains separate reconstruction and motion branches, injecting attention maps or value features into the main branch.
Although these algorithms do not require the construction of data or training models, the quality of the edits often falls short in terms of temporal consistency, text alignment, and background preservation, among other factors.

\subsection{Training-based Video Editing}
Training-based approahces~\cite{wu2023tuneavideo,tu2024motioneditor,qin2024instructvid2vid,cheng2023consistent,wang2024videocomposer,zhang2024avid,zi2024cococo} often achieve better editing quality.
Several algorithms~\cite{wu2023tuneavideo,tu2024motioneditor} extend the text-to-image model to a text-to-video model, employing one-shot learning, employing one-shot learning to extract motion information into the model parameters.
Other algorithms~\cite{qin2024instructvid2vid,cheng2023consistent} generate synthetic datasets based on training-free or one-shot approaches, which are then used to train models. 
However, the editing quality is constrained by the generation quality of the data generator.
Recently, video local editing algorithms~\cite{wang2024videocomposer,zhang2024avid,zi2024cococo} introduce automated data construction pipelines and train the model on real-world data. 
These algorithms mask entities in videos and use LLMs to generate local captions for the masked regions. 
The masked video serves as the model input, while the original video is used as the ground truth during training. 
Video inpainting models~\cite{zhou2023propainter,li2025diffueraser} are trained by adding random mask sequences to simulate the entity deletion and restore the video content.
\begin{figure*}[t]
    \centering
    \includegraphics[width=\linewidth]{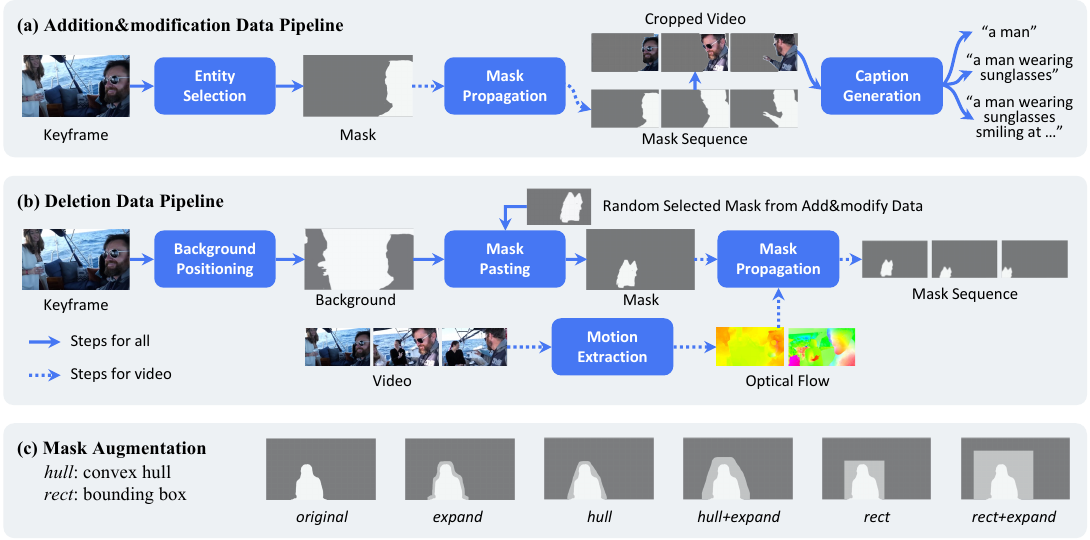}
    \caption{
    Data construction pipelines, where solid lines are required for both image and video data, and the dashed lines are only for video.
    }
    \label{fig:pipe}
\end{figure*}

\section{VIVID-10M Dataset}\label{sec:dataset}
In this section, we introduce VIVID-10M, which, to the best of our knowledge, is the first open-source large-scale video local editing dataset. It covers a range of tasks including addition, modification, and deletion (\cref{fig:dataset_vis}).
Each training sample is a tuple $(x,m,\tilde{x},y)$, where $x=\{x^i\}$ represents a video or image. $i$ denotes the $i$-th frame of the video, and we consider an image as a video with only one frame. $m=\{m^i\}$ denotes the corresponding binary masks of the editing area, $\tilde{x}=\{\tilde{x}^i\}$ is the masked video or image, and $y$ is a caption of the editing area.
In the masked video or image, the editing regions are erased, while the non-editing regions are preserved, so $\tilde{x}^i=x^i\odot (1-m^i)$.

VIVID-10M contains two subsets, VIVID-10M-Video and VIVID-10M-Image, both derived from the publicly available PANDA-70M dataset~\cite{chen2024panda70m}.
The video subset includes 73.7K videos, each at least 5 seconds in length. 
The image subset contains the first frame extracted from 672.7K videos.
Subsequent sections detail the dataset construction methods for various tasks (\cref{sec:modify_pipe} and \cref{sec:remove_pipe}).
We also proposed a data augmentation method in \cref{sec:data_augmentation} ,
designed to diversify the mask into six distinct types, varying in shape and scale.
Finally, we provides statistics in \cref{sec:statics}.

To accurately evaluate the editing performance of the model, we manually construct a high-quality validation dataset, VIVID-10M-Eval (detailed in Appendix).

\subsection{Addition\&Modification Data Pipeline}\label{sec:modify_pipe}
The addition task adds new entities to the video, while the modification task changes the type or attributes of existing entities.
The goal of both tasks is to draw entities in the mask area within the video.
To unify the training data formats for both tasks, we select entities from images and videos and generate corresponding local captions. 
As shown in \cref{fig:pipe}(a), the pipeline for VIVID-10M-Video consists of three stages: entity selection, mask propagation, and local caption generation, while the pipeline for VIVID-10M-Image omits the mask propagation stage.

\noindent{\textbf{Entity Selection.}}
In this stage, editable entities are selected from the image or the first frame of the video, followed by mask generation.
Specifically, we first apply RAM~\cite{zhang2024ram} to extract entity labels from the frame and filter the labels using a predefined vocabulary (see Appendix).
Then, we use Grounding DINO~\cite{liu2023groundingdino} to detect the bounding boxes corresponding to the labels.
Finally, each box serves as a prompt for SAM2~\cite{SAM2} to generate the mask.

\noindent{\textbf{Mask Propagation.}}
For video data, the editable area must track the movement of the entity across the frames.
Therefore, we use SAM2~\cite{SAM2} to propagate the mask from the first frame to the subsequent frames.

\noindent{\textbf{Local Caption Generation}}
In this stage, we generate local captions for the editing areas.
First, we use $x$ and $m$ to crop entities from the video or image $\{x^i\odot m^i\}$, where the non-editing areas are erased.
These cropped inputs, denoted as $\hat{x}$ are then fed into InternVL2~\cite{chen2024internvl} to generate local captions of three different lengths. 
The prompt we used for InternVL2 is detailed in the Appendix.

\begin{table*}[t]
    \centering
    \small
    \begin{tabular}{lcccccccc}
    \toprule
    Dataset &\#Addition\&Modification & \#Deletion & \#Image/Video & \#Entity Label & MG & MP & TA & HQ \\
    \midrule
    VIVID-10M-Image & 7,578,102 & 671,837 & 672,715 & 8,237 & 0.83 & - & 0.86 & 0.74 \\
    VIVID-10M-Video & 1,313,906 & 204,434 & 73,737 & 2,962 & 0.81 & 0.72 & 0.86 & 0.60 \\
    \bottomrule
    \end{tabular}
    \caption{Statistics. High-quality (HQ) data is assessed via Mask Generation (MG), Mask Propagation (MP), and Text Alignment (TA).}
    \label{tab:high_quality_data}
\end{table*}

\subsection{Deletion Data Pipeline}\label{sec:remove_pipe}
The deletion task involves removing existing entities from the video and inpainting these areas with background pixels. 
Unlike the other two tasks, paired data for the deletion task cannot be generated simply by masking existing entities, as this task requires ground truth background pixels for effective training. 
To address this, we construct the deletion dataset by adding entity masks from other videos to the background areas.
\cref{fig:pipe}(b) illustrates the pipeline of the deletion task.
The deletion pipeline consists of three stages: background positioning, mask pasting, and mask propagation.
The local caption for deletion task is fixed: \textit{``Remove objects and generate areas that blend with the background."}

\noindent{\textbf{Background Positioning.}}
Similar to the \textit{Entity Selection} stage in the addition\&modification pipeline, we use RAM~\cite{zhang2024ram}, Grounding DINO~\cite{liu2023groundingdino} and SAM2~\cite{SAM2} to identify the background areas in the first frame.
The vocabulary is replaced with the background vocabulary (see Appendix).

\noindent{\textbf{Mask Pasting.}}
To align with inference, we paste  entity masks from other videos to the background areas.
Specifically, we randomly select a mask sequence from the addition\&modification samples and paste the first mask into the background area of the keyframe.

\noindent{\textbf{Mask Propagation.}}
There are two possible scenarios for the deletion task: 1) deleting a foreground entity (\textit{e.g.}, removing a running car) and 2) deleting a background entity (\textit{e.g.}, removing a picture frame from the wall).
In the first case, the entity follows its trajectory, so we directly copy the subsequent masks and paste them into the subsequent frames.
In the second case, the entity's trajectory is aligned with the background.
Therefore, we use RAFT~\cite{teed2020raft} to calculate the optical flow of the background pixels and propagate the mask on keyframe to the subsequent frames.

\subsection{Mask Augmentation}\label{sec:data_augmentation}
The pipelines described in \cref{sec:modify_pipe} and \cref{sec:remove_pipe} generate masks that strictly match the entity shapes, which may leak semantic information and reduce the robustness of the editing model.
To address this issue and expand the dataset, we apply data augmentation.
Three operators are employed: \textit{expand}, \textit{hull}, and \textit{box}.
The \textit{expand} operator randomly enlarges the mask while preserving its original shape. The \textit{hull} operator calculates the convex hull of the mask, and the \textit{box} operator determines the bounding box.
By combining these operators, we derive five new masks:
1) \textit{expand}, 2) \textit{hull}, 3) \textit{box}, 4) \textit{hull+expand}, 5) \textit{box+expand} (\cref{fig:pipe}(c)).

\begin{figure}[t]
    \centering
    \begin{minipage}[b]{0.47\linewidth}
        \centering
        \includegraphics[width=\linewidth]{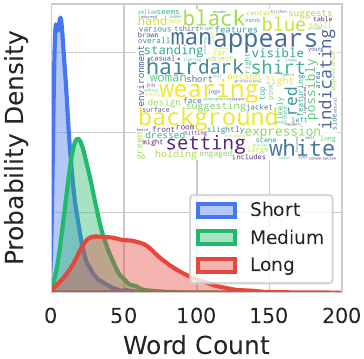}
        \caption{Caption distribution. Each sample has three captions of different lengths.}
        \label{fig:caption_stats}
    \end{minipage} \hfill
    \begin{minipage}[b]{0.5\linewidth}
        \centering
        \includegraphics[width=\linewidth]{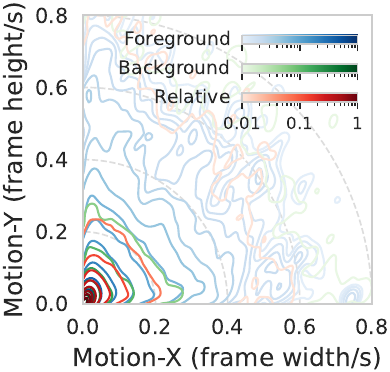}
        \caption{Motion distribution. Foreground is editing area, background is non-editing area.}
        \label{fig:flow_stats}
    \end{minipage}
\end{figure}

\subsection{Statistics}\label{sec:statics}
The statistics of VIVID-10M are shown in \cref{tab:high_quality_data}.
We apply filters for each component of the pipeline to ensure the data quality (see Appendix).
To evaluate the quality of the datasets, we measure the quality from three dimensions using user study: Mask Generation (MG), Mask Propagation (MP) and Text Alignment (TA).
\cref{tab:high_quality_data} shows that the two subsets perform similarly in terms of MG and TA metrics, while VIVID-10M-Video introduces additional noise in the MP process, which ultimately leads to lower ratio of high-quality data (HQ) than VIVID-10M-Image.
This demonstrates that using image data to expand video data can effectively reduce the construction cost of high-quality data.
As shown in \cref{fig:caption_stats}, VIVID-10M covers captions of various lengths and contains rich semantics.
\cref{fig:flow_stats} illustrates the motion distribution of mask area (foreground), background and relative motion between them, which shows VIVID-10M contains data for various movement intensity.

\begin{figure}[t]
    \centering
    \includegraphics[width=\linewidth]{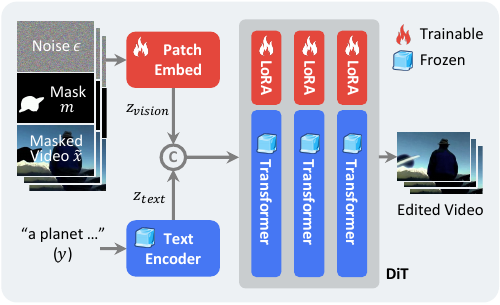}
    \caption{Model architecture of VIVID.}
    \label{fig:model}
\end{figure}
\section{VIVID Model}
To validate VIVID-10M, this section outlines a versatile and interactive video local editing model. 
Specifically, \cref{sec:pre} covers the foundational principles, \cref{sec:model} introduces the VIVID architecture, \cref{sec:easyedit} presents the keyframe-guided interactive video editing for efficient video editing, and \cref{sec:train} discusses our multi-task joint training. 

\subsection{Preliminaries}\label{sec:pre}
Video editing model can be framed as a conditional diffusion model, where the model $\epsilon_\theta$ is trained to predict noise based on given conditional information.
The optimization objective of the video editing model is defined as \cref{eq:condition}.
\begin{equation}
    \mathcal{L}(\theta)=\mathbb{E}_{\epsilon \sim \mathcal{N}(0,\mathrm{I})}[\|\epsilon-\epsilon_{\theta}(x_{t},t,c)\|_{2}^2],
\label{eq:condition}
\end{equation}
where $t\in\{1,...,T\}$ represents the number of diffusion steps, $x_t$ denotes the noised video, and $c$ represents the conditional inputs (\textit{e.g.}, the caption and masks).


\subsection{Architecture}\label{sec:model}
We introduce VIVID, a versatile and interactive video editing model, that supports adding, modifying, and deleting entities within specific region.
Given a video $x$, VIVID generates high quality, harmonious contents within the mask sequence $m$, guided by the semantics of local caption embedding $\tau_{\theta}(y)$.
Using the optimization target defined in \cref{eq:condition}, we set the condition $c=(\tilde{x},m,\tau_{\theta}(y))$.
VIVID builds upon CogVideoX~\cite{yang2024cogvideox} to leverage its pretrained video generation capabilities.
\cref{fig:model} highlights the trainable components, including LoRA~\cite{hu2021lora} and the patch encoder.
Specifically, we concatenate the mask sequence $m$ and the masked video $\tilde{x}$ with the noise, converting them into visual latent $z_{vision}$.
Since the input dimensions of the patch embedder are different with text-to-video generation~\cite{yang2024cogvideox}, it is also trained.
Meanwhile, we obtain the textual latent $z_{text}=\tau_{\theta}(y)$ from the local caption $y$ using a text encoder $\tau_{\theta}$.
Finally, $z_{vision}$ and $z_{text}$ are concatenated and input to the DiT to generate the edited video.

\begin{figure}[t]
    \centering
    \includegraphics[width=\linewidth]{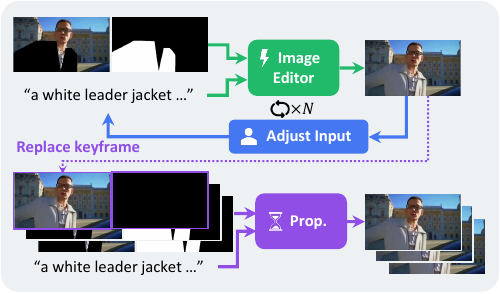}
    \caption{Keyframe-guided interactive video editing mechanism, where \textit{Prop.} means using VIVID to propagate the edit of the keyframe to remaining frames.}
    \label{fig:easy_edit}
\end{figure}
\subsection{Keyframe-guided Interactive Video Editing}\label{sec:easyedit}
In practical video editing scenarios, users often cannot fully express their requirements in a single attempt, leading to iterative adjustments of the local caption based on model feedback.
This process requires multiple model runs to achieve satisfactory results, increasing both time and resource demands and potentially compromising user experience.
To address this, we propose the Keyframe-guided Interactive Video Editing (KIVE) mechanism, shown in \cref{fig:easy_edit}, which enables users to quickly edit keyframes using an image editing model and propagate these edits across the remaining frames. 
Assume that we have both an image editing model and a video editing model with comparable generative capabilities and respective inference costs $c_{im}$ and $c_{vid}$.
If users need an average of $N$ edits to reach a satisfactory result, the cost for direct video editing would be $N\cdot c_{vid}$, whereas the cost using KIVE is only $N\cdot c_{im}+c_{vid}$.
As $N$ grows, the advantages become pronounced.
To enable VIVID to support KIVE, we train it by replacing the first frame of masked video with the original video, and the first mask with the all-black frame $\mathbf{0}$.
Thus, the conditional input can be represented as $\bar{c}=({\bar{x},\bar{m},\tau_{\theta}(y)})$,
where $\bar{x}=\{x_0^0\}\cup\{\tilde{x}_0^{i}\}_{i>0}$ and $\bar{m}=\{\mathbf{0}\}\cup\{m^{i}\}_{i>0}$ represent the masked video and mask sequence with the first frame replaced.
Additionally, selecting the last frame of an edited clip as the keyframe for the next enables VIVID to edit long videos of through the KIVE mechanism (see Appendix).

\begin{figure*}[t]
    \centering
    \includegraphics[width=\linewidth]{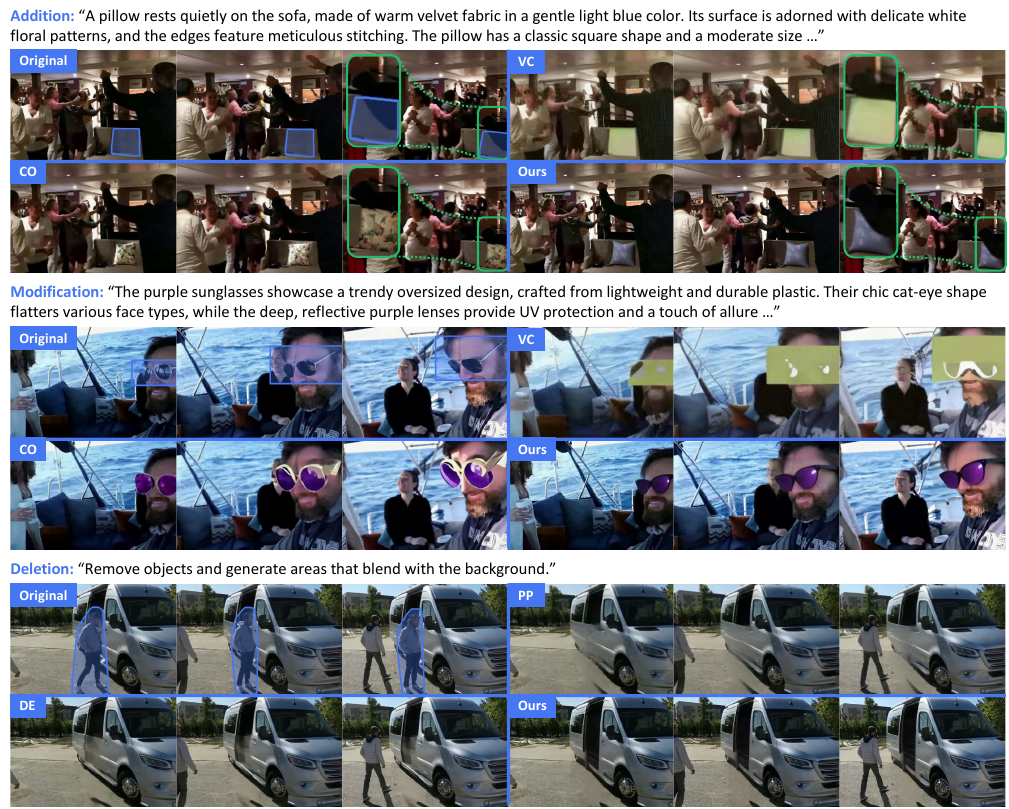}
    \caption{The editing results for VIVID (Ours), VideoComposer (VC), COCOCO (CO), ProPainter (PP) and DiffuEraser (DE).}
    \label{fig:compare_methods}
\end{figure*}
\subsection{Multi-Task Joint Training}\label{sec:train}
To reduce training overhead and accelerate convergence, we incorporate both image and video data during training.
As noted in \cref{sec:statics}, image data offers greater diversity and a higher proportion of high-quality samples.
Given a fixed training time, leveraging this broader image dataset improves the model's generalization capability for edits.
Our default configuration uses an image-to-video ratio of 10:1.
At each training step, we set the batch to consist entirely of either images or videos based on this proportion, maximizing training efficiency.
In addition, to support the KIVE mechanism, we randomly replace the conditional input $c$ in video editing with $\bar{c}$ with 50\% probability during training.
Recognizing that addition and modification tasks are more challenging than deletion task, since they require generating clear foreground information, we adjust the data ratio of different tasks to addition\&modification:deletion=3:1.
\begin{table*}[t]
    \centering
    \small
    \begin{minipage}[b]{0.6\linewidth}
    \centering
        \begin{tabular}{llccccccc} \toprule
        \multirow{2}{*}{Tasks} & \multirow{2}{*}{Models} & \multicolumn{3}{c}{Automated Metrics} & \multicolumn{4}{c}{User Study (Win Rates\%)}  \\ \cline{3-9}
        & & TC$\uparrow$ & TA$\uparrow$ & BP$\downarrow$ & VQ$\uparrow$ & TC$\uparrow$ & TA$\uparrow$ & BP$\uparrow$ \\ \midrule
        \multirow{3}{*}{Add.}& VC~\cite{wang2024videocomposer} & 95.61 & 18.09 & 77.23 & 1.19 & 4.76 & 15.61 & 0.79 \\
        & CO~\cite{zi2024cococo} & \textbf{97.17} & 20.67 & 25.82 & 52.38 & 55.69 & 54.10 & 95.24 \\
        & \textbf{Ours} & 96.34$^{\dagger}$ & \textbf{21.45} & \textbf{18.17} & \textbf{83.86} & \textbf{93.25} & \textbf{67.06} & \textbf{99.73} \\
        \midrule
        \multirow{3}{*}{Mod.}& VC~\cite{wang2024videocomposer} & 95.02 & 19.48 & 59.75 & 0.83 & 6.21 & 6.04 & 1.14 \\
        & CO~\cite{zi2024cococo} & \textbf{96.83} & 22.31 & 20.62 & 49.85 & 44.16 & 49.67 & 94.42 \\
        & \textbf{Ours} & 96.14$^{\dagger}$  & \textbf{23.56} & \textbf{14.84} & \textbf{83.30} & \textbf{96.88} & \textbf{79.06} & \textbf{98.76} \\
        \midrule
        \multirow{3}{*}{Del.}& PP~\cite{zhou2023propainter} & \textbf{98.97} &-& \textbf{13.86} & 18.93 & 20.67 & 45.45 & 55.52 \\
        & DE~\cite{li2025diffueraser} & 98.88 &-& 14.36 & 62.53 & 45.86 & 82.84 & 79.38 \\
        & \textbf{Ours} & 98.78  &-& 18.41 & \textbf{67.84} & \textbf{85.89} & \textbf{86.18} & \textbf{91.81} \\
        \bottomrule
        \end{tabular}
        \caption{
        Comparison of automated metrics and user studies (win-or-draw rate) for VIVID (Ours), VideoComposer (VC), COCOCO (CO), ProPainter (PP), and DiffuEraser (DE). $^{\dagger}$Results after downsample to 7.5fps.
        }
        \label{tab:method_compare}
    \end{minipage}
    \hspace{4mm}
    \begin{minipage}[b]{0.36\linewidth}
    \begin{minipage}[b]{\linewidth}
        \centering
        \begin{tabular}{lcccc} \toprule
        Algorithms & TC$\uparrow$ & TA$\uparrow$ & BP$\downarrow$ \\ \midrule
        Video Editing & 96.26 & 22.51 & 17.14 \\
        KIVE & 96.21 & 22.11 & 17.07 \\
        \bottomrule
        \end{tabular}
        \caption{Automated metrics for video editing and KIVE mechanism.}
        \label{tab:easy_edit}
    \end{minipage}
    
    \begin{minipage}[b]{\linewidth}
        \vspace{3.7mm}
        \centering
        \begin{tabular}{lcccc} \toprule
        Settings & TC$\uparrow$ & TA$\uparrow$ & BP$\downarrow$ \\
        \midrule
        Image:Video=0:1 & 96.03 & 21.72 & 31.47  \\
        Image:Video=1:1 & \textbf{96.14} & 21.71 & 19.63  \\
        Image:Video=5:1 & 96.14 & 21.61 & 20.88 \\
        \textbf{Image:Video=10:1} & 95.87 & \textbf{22.11} & \textbf{19.18} \\ 
        \bottomrule
        \end{tabular}
        \caption{Automated metrics for various ratios of image and video joint training.}
        \label{tab:joint_train}
    \end{minipage}
    \end{minipage}
\end{table*}
\section{Experiments}
\subsection{Setup}
\textbf{Implementation details.} 
Our approach builds on the CogVideoX 5B model~\cite{yang2024cogvideox}. 
We train VIVID on VIVID-10M using LoRA~\cite{hu2021lora} at a resolution of $480 \times 720$ for the original video frames, with a LoRA rank of 32. 


\noindent\textbf{Baselines.} 
We evaluate VIVID on VIVID-10M-Eval, which comprises three editing tasks: \textit{1) addition, 2) modification,} and \textit{3) deletion}.
For the addition and modification tasks, we select VideoComposer~\cite{wang2024videocomposer} and COCOCO~\cite{zi2024cococo} as the baseline models.
We modify the local caption for VideoComposer to a global caption to match its training setup.
For the deletion task, ProPainter~\cite{zhou2023propainter} and DiffuEraser~\cite{li2025diffueraser} are introduced as the baseline models.

\noindent\textbf{Evaluation Metrics.} 
\textit{(a) Automatic Metric Evaluation}. 
Background Preservation (BP): the L1 distance between the original and edited videos in non-editing regions. 
Text Alignment (TA): the CLIP-score~\cite{zhang2024avid, hessel2023clipscore} of the edited region. 
Temporal Consistency (TC): the cosine similarity between consecutive frames in the CLIP-Image feature space~\cite{zhang2024avid}. 
\textit{(b) User study}. 
To better align with human perception, we also conduct a user study, where annotators evaluate the edits across BP, TA, TC and Visual Quality (VQ), where VQ reflects the realness and aesthetics of videos.
The final results are presented as win rates.


\subsection{Comparisons}
\noindent\textbf{Qualitative comparisons.}
We provide editing examples of VIVID and baseline models~\cite{zi2024cococo,wang2024videocomposer,zhou2023propainter,li2025diffueraser} in \cref{fig:compare_methods}.
Our approach achieves more aesthetic and semantically correct edits across all tasks.
For example, in the addition task, VIVID edits the correct color according to the text caption and inpaints the arm in the last frame, demonstrating an understanding of occlusion.
In the modification task, VIVID's edited sunglasses maintain a consistent structure and appearance as the man's head turns.
Finally, in the deletion task, VIVID effectively inpaints the open car door.
More qualitative comparison are shown in the Appendix.
We also provide a \textbf{\textit{Demo page}} in the supplementary files, to better exhibit our versatile and aesthetic edits.

\noindent\textbf{Quantitative comparisons.}\label{sec:quantitative_exp}
Quantitative results in \cref{tab:method_compare} show that VIVID achieves better or comparable performance in automated metrics compared to other models.
Specifically, for TC, VIVID surpasses VideoComposer~\cite{wang2024videocomposer}, and performs similar to other methods~\cite{zi2024cococo,zhou2023propainter,li2025diffueraser}.
It is worth noting that, we downsample frames to 7.5fps for the addition and modification tasks, matching other models, which reduces VIVID's TC (see Appendix).
For TA, we only measure the performance of the addition and modification tasks due to the fixed caption in deletion (\cref{sec:remove_pipe}).
VIVID leads in both tasks, indicating strong caption consistency.
Finally, VIVID achieves lower BP value to VideoComposer~\cite{wang2024videocomposer} and COCOC~\cite{zi2024cococo}, and is comparable with ProPainter~\cite{zhou2023propainter} and DiffuEraser~\cite{li2025diffueraser}.

\noindent\textbf{User Study}
Compared to automated metrics, user studies provide more meaningful insights~\cite{polyak2024moviegen}.
\cref{tab:method_compare} show that our model has substantially higher scores in VQ, TA, and TC across all tasks, indicating VIVID's ability to produce aesthetically pleasing, semantically aligned, and temporally stable edits.
For BP, VIVID performs comparably to baseline methods~\cite{zi2024cococo,li2025diffueraser}.
The discrepancy between user studies and automated metrics in the TC is because automated metrics only capture semantic changes between frames, overlooking pixel level instability. 
Our model shows significantly reduced jitter and flicker.

\begin{figure}[t]
    \centering
    \includegraphics[width=\linewidth]{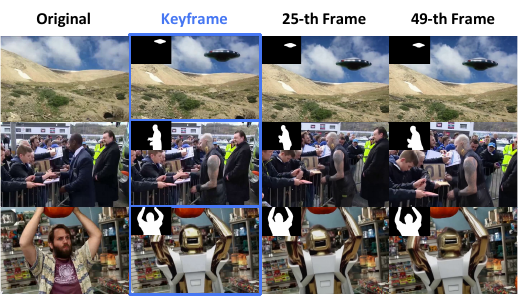}
    \caption{The editing results of KIVE mechanism.}
    \label{fig:easy_edit_result}
\end{figure}

\subsection{Effectiveness of KIVE}\label{sec:exp_easyedit}
As shown in \cref{tab:easy_edit}, the editing quality achieved with the KIVE mechanism is comparable to that of direct video editing.
Examples of keyframe image editing and propagation are also displayed in \cref{fig:easy_edit_result}, demonstrating that VIVID not only perform high-quality image editing but also preserves entity features in subsequent frames.
Additionally, VIVID consumes 17.1 peta floating-point operations (PFLOPs) for video, while only 1.5 PFLOPs for keyframe editing.
This reduction highlights KIVE's efficiency, enabling users to interactively refine local captions and achieve satisfying high-quality results more effectively.

\subsection{Ablation Study}

\noindent\textbf{Mixture of Image and Video Data.}
We evaluate the impact of varying image-to-video ratios on editing quality, comparing ratios of 10:1, 5:1, 1:1 and 0:1 under identical training time.
\cref{tab:joint_train} shows that integrating image data with video data effectively enhances TA and BP without seriously compromising TC.
Notably, a 1:1 ratio already lowers BP score from 31.47 to 19.63.
Since the 10:1 image-to-video ratio yields the best TA and BP, and maintains TC comparable to other settings, making it our default setting.

\begin{figure}[t]
    \centering
    \begin{minipage}[b]{0.5\linewidth}
        \centering
        \includegraphics[width=\linewidth]{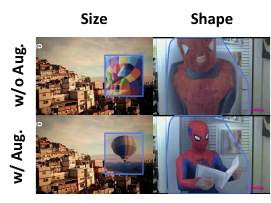}
        \caption{Editing results with and without data augmentation.}
        \label{fig:data_aug}
    \end{minipage} \hfill
    \begin{minipage}[b]{0.45\linewidth}
    \setlength{\tabcolsep}{3pt} 
    \makeatletter\def\@captype{table}
        \centering
        \small
        \begin{tabular}{cccc}
        \toprule
        & Tasks & Size$\uparrow$ & Shape$\uparrow$ \\
        \midrule
        \multirow{2}{*}{w/o} & Add. & 40.48 & 35.37 \\
        & Mod. & 49.21 & 34.76 \\
        \midrule
        \multirow{2}{*}{\textbf{w/}} & \textbf{Add.} & \textbf{87.30} & \textbf{93.90} \\
        & \textbf{Mod.} & \textbf{84.13} & \textbf{92.07} \\
        \bottomrule
        \end{tabular}
        \caption{User study results with and without data augmentation (win‐or‐draw rate).}
        \label{tab:data_aug}
    \end{minipage}
\end{figure}

\noindent\textbf{Data Augmentation.}
To examine the effects of data augmentation on editing quality, we evaluated edits using records from the VIVID-10M dataset that do not include augmented masks (\cref{sec:data_augmentation}).
We compared models trained with and without augmented data, each on 832K samples.
As shown in \cref{fig:data_aug}, training the model on augmented data effectively alleviates overfilling and entity deformity issues, enabling it to edit entities that differ in shape and scale from the mask.
We also conduct a user study (\cref{tab:data_aug}) and find that the model with augmentation achieves significantly higher win‐or‐draw rates.
This enhances users experience by reducing the need for exact mask inputs.
\section{Conclusion}
We introduce VIVID-10M, the first large-scale video local editing dataset created to overcome the high costs of constructing paired datasets and training models for video local editing.
Leveraging VIVID-10M, our proposed VIVID model demonstrates strong performance in addition, modification, and deletion tasks. 
The introduction of the keyframe-guided interactive video editing mechanism enhances user interaction by enabling iterative keyframe adjustments and efficient propagation of edits across frames, significantly reducing latency in achieving satisfactory results. 
Experimental results confirm that VIVID achieves state-of-the-art performance, surpassing existing models in both automated metrics and user studies. 

\noindent\textbf{Limitations.}
VIVID faces challenge when it comes to more global editing tasks (\textit{e.g.}, stylization) or more fine-grained editing tasks (\textit{e.g.}, attribute editing).
We will explore more editing tasks in future work.

{
    \small
    \bibliographystyle{ieeenat_fullname}
    \bibliography{main}
}


\end{document}